\documentclass[sigconf]{acmart}

\copyrightyear{2024}
\acmYear{2024}
\setcopyright{rightsretained}
\acmConference[HRI '24]{Proceedings of the 2024 ACM/IEEE International Conference on Human-Robot Interaction}{March 11--14, 2024}{Boulder, CO, USA}
\acmBooktitle{Proceedings of the 2024 ACM/IEEE International Conference on Human-Robot Interaction (HRI '24), March 11--14, 2024, Boulder, CO, USA}
\acmDOI{10.1145/3610977.3634973}
\acmISBN{979-8-4007-0322-5/24/03}

\makeatletter
\gdef\@copyrightpermission{
  \begin{minipage}{0.3\columnwidth}
    \href{https://creativecommons.org/licenses/by/4.0/}{\includegraphics[width=\textwidth]{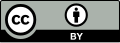}}
  \end{minipage}
  \hfill
  \begin{minipage}{0.7\columnwidth}
    \href{https://creativecommons.org/licenses/by/4.0/}{This work is licensed under a Creative Commons Attribution International 4.0 License.}
  \end{minipage}
  \vspace{5pt}
}
\makeatother

\AtBeginDocument{%
  \providecommand\BibTeX{{%
    \normalfont B\kern-0.5em{\scshape i\kern-0.25em b}\kern-0.8em\TeX}}}

\vspace{-5mm}

\begin{document}
\setlength{\abovedisplayskip}{1pt}
\setlength{\belowdisplayskip}{1pt}
\title{Improving Explainable Object-induced Model through Uncertainty for Automated Vehicles}


\author{Shihong Ling}
\affiliation{%
  \institution{School of Computing and Information\\University of Pittsburgh}
  \city{Pittsburgh}
  \state{PA}
  \country{USA}
}
\email{shl282@pitt.edu}

\author{Yue Wan}
\affiliation{%
  \institution{School of Computing and Information\\University of Pittsburgh}
  \city{Pittsburgh}
  \state{PA}
  \country{USA}
}
\email{yuw253@pitt.edu}

\author{Xiaowei Jia}
\affiliation{%
  \institution{School of Computing and Information\\University of Pittsburgh}
  \city{Pittsburgh}
  \state{PA}
  \country{USA}
}
\email{xiaowei@pitt.edu}

\author{Na Du}
\affiliation{%
  \institution{School of Computing and Information\\University of Pittsburgh}
  \city{Pittsburgh}
  \state{PA}
  \country{USA}
}
\email{na.du@pitt.edu}


\begin{abstract}
The rapid evolution of automated vehicles (AVs) has the potential to provide safer, more efficient, and comfortable travel options. However, these systems face challenges regarding reliability in complex driving scenarios. Recent explainable AV architectures neglect crucial information related to inherent uncertainties while providing explanations for actions. 
To overcome such challenges, our study builds upon the "object-induced" model approach that prioritizes the role of objects in scenes for decision-making and integrates uncertainty assessment into the decision-making process using an evidential deep learning paradigm with a Beta prior. Additionally, we explore several advanced training strategies guided by uncertainty, including uncertainty-guided data reweighting and augmentation. Leveraging the BDD-OIA dataset, our findings underscore that the model, through these enhancements, not only offers a clearer comprehension of AV decisions and their underlying reasoning but also surpasses existing baselines across a broad range of scenarios.
\end{abstract}




\maketitle

\section{Introduction}
Advancements in artificial intelligence have led to the proliferation of automated vehicles (AVs), offering the potential for safer and more efficient intelligent transportation systems. 
However, the transformative nature of AVs is hindered by a critical issue: the lack of system transparency. This opacity leads drivers to view AVs as black boxes, causing unwarranted interventions or oversights in emergent situations that require prompt responses \cite{rudin2019stop}.

There are two major approaches to developing AVs: the end-to-end manner, which directly connects sensory input to driving actions \citep{xu2017end, wang2019deep}, and pipeline architecture, which processes intermediate stages before decision-making. While end-to-end systems excel in using comprehensive visual information for decision-making, their complex architectures impede explainability. Conversely, pipeline approaches, though more interpretable, suffer from error propagation and limited performance in diverse scenarios. This dichotomy highlights the urgent need for AV models that balance performance with transparency \cite{koo2015kwac, kim2017interpretable, kim2018textual, xu2020explainable, omeiza2021towards}.

One noteworthy idea is the "object-induced" model, which is inspired by how humans solve the problem. When deciding among driving actions such as forwarding and turning, humans do not employ a strict end-to-end strategy. Instead, they perform a certain amount of understanding and reasoning about scene objects. Similarly, this idea takes into account the importance of objects and their roles within a scene in making driving decisions. Notably, the work of \cite{kim2018textual} and \cite{xu2020explainable} exemplifies this trajectory. They utilized convolutional neural networks (CNNs) to establish correlations between visual inputs and vehicle decisions while emphasizing important objects in scenes. These methodologies were further enhanced through attention-based video-to-text models and advanced CNN architectures, whose performance was validated using the Berkeley DeepDrive eXplanation (BDD-X) dataset \cite{yu2020bdd100k}.

Despite these advancements deepening our understanding of AV systems, an important challenge remains: ensuring the transparency and reliability of AVs in complex and unpredictable scenarios. Regardless of their foundational design, many existing AV models offer explanations and advice with an unacknowledged level of confidence. This becomes problematic when these systems provide potentially misguided instructions without conveying their inherent uncertainty, thereby posing the potential risk of causing severe safety issues during autonomous driving. For instance, consider a nighttime driving situation with reduced visibility and a pedestrian crossing the road. In this scenario, the AV, without considering its own uncertainty, may make a deterministic decision to proceed, failing to recognize the complexities and its lack of confidence in the decision. Neglecting this uncertainty, the AV may mistakenly perceive the situation as offering a clear and confident decision path. This example exemplifies how real-world conditions are often characterized by ambiguity. In the absence of a mechanism for addressing inherent uncertainties, AV systems have trouble effectively handling unexpected cases.

Thus, we aim to enhance AV's explanation generation by leveraging the uncertainty information in taking actions. 
Instead of solely relying on deterministic explanations, we argue that assessing the uncertainty within model predictions can provide drivers with a reliable perspective of the model's level of confidence, reasoning, and potential limitations. 

By utilizing the BDD-OIA dataset \cite{xu2020explainable}, this paper aims to address the ambiguous reasoning behind AV actions and explanations. The underlying deep learning architecture has been modified to account for specific driving actions while incorporating uncertainty as a fundamental aspect. The proposed methodology follows the evidential deep learning paradigm that uses a Beta prior to capture and encapsulate both model and data uncertainties. As demonstrated in our study, these integrations effectively improve the reliability of the model.

To sum up, our contributions are threefold:
\vspace{-1mm}
\begin{itemize}
\item We have refined the explainable object-induced model.
\item We introduce the uncertainty-guided training strategies that improve over baseline methods by a large margin, thereby demonstrating their effectiveness.
\item Case study shows that our method can enhance model interpretability in challenging driving scenarios.
\end{itemize}

\vspace{-1mm}
\section{Related Work}
\vspace{-1mm}
\subsection{Explainable Autonomous Driving System} 
\vspace{-1mm}
\subsubsection{From Modular Pipelines to End-to-End Learning.}

Rapid development in computing algorithms and machine learning techniques have greatly enhanced autonomous driving systems over the past few decades. Modular pipelines, which break down the driving task into sub-problems like perception and planning, were the initial focus due to their somewhat explainable nature \citep{zablocki2022explainability}. However, their reliance on human heuristics and manual intermediate representations limits their adaptability to real-world uncertainties. As such, attention has turned to end-to-end learning models that map sensor data directly to driving actions \citep{kim2017interpretable, xu2017end, kim2018textual}. These models have grown more accurate with advances in computer vision. Nonetheless, achieving improved accuracy alone is insufficient for AV systems. The reliability in real-world scenarios, including both common and adversarial situations, as well as the transparency of the decision-making process, are much more critical. These aspects pose unavoidable challenges for end-to-end learning models, which are typically trained in simulated environments and produce unexplainable decision-making.
\vspace{-1mm}
\subsubsection{Enhancing Reasoning for Autonomous Driving.}

To enhance the reasoning capabilities of autonomous driving systems and their real-world reliability, previous studies have conducted extensive investigations on evaluating the effectiveness of diverse reasoning techniques and developing reasoning AV models. For instance, \citet{colley2021effects} evaluated the effects of semantic segmentation visualization technique on trust, situation awareness, and cognitive load which was later expanded into other visualization techniques in \cite{colley2022effects}. \citet{shen2022explain2} and \citet{omeiza2021not} discussed the circumstances under which explanations are necessary and how the content of these explanations changes in the context of autonomous driving. \citet{du2019look} examined explanation timing and found that explanations provided before an AV acted unexpected behaviors were associated with higher trust in and preference for AVs. 

Recent efforts have further enriched the explainability landscape in autonomous driving. \citet{cultrera2020explaining} proposed to train an imitation learning-based agent equipped with an attention model. \citet{xu2020explainable} developed an explainable object-induced AV system that emulates human decision-making processes. \citet{ben2022driving} introduced the BEEF architecture, providing high-level driving explanations by fusing features to elucidate the behavior of a trajectory prediction model. \citet{atakishiyev2023explain2} presented a Visual Question Answering (VQA) framework for interpretable decision-making in autonomous driving, employing question-answer pairs as justifications for actions. Among these works, the work \cite{xu2020explainable}, which integrates object-specific reasoning within a broader scene, stands out for its potential in improving action predictions and explanations in AVs. This framework serves as the foundation for our research, which aims to further enhance the reasoning strategies of autonomous driving systems.

\vspace{-1mm}
\subsection{Uncertainty Quantification}
Uncertainty plays a fundamental role in decision-making across various fields. In the context of deep learning, accurately representing uncertainty is critical for ensuring reliable predictions \cite{jiang2018trust}. Machine learning uncertainties can be broadly classified into two categories: aleatoric and epistemic. Aleatoric uncertainty refers to the intrinsic data randomness that cannot be explained. Epistemic uncertainty, also known as model uncertainty, arises due to gaps between our knowledge and model designs \cite{hullermeier2021aleatoric}. Previous researchers mainly explored two types of uncertainty quantification methods: Bayesian approximation and ensemble learning techniques. The Bayesian method offers a principled probabilistic framework for uncertainty quantification, deriving uncertainty directly from the posterior distributions of model parameters \citep{gal2016dropout, kupinski2003ideal, swiatkowski2020k}, while the ensemble method derives uncertainty from the variation in predictions across multiple models \citep{hu2019mbpep, fersini2014sentiment}. 

Additionally, instead of using model ensembles to capture output variation, a group of methods incorporate the Dirichlet distribution to model the predictive distributions. For example, \citet{tsiligkaridis2021information} proposed the Information Aware Dirichlet networks, which refine the prediction confidence by leveraging a Dirichlet prior distribution on predictions, differentiating it from conventional neural networks. Furthermore, \citet{sensoy2018evidential} proposed Evidential Deep Learning (EDL) to capture the output uncertainty, treating neural network predictions as subjective opinions and employing a Dirichlet distribution for class probabilities, highlighting the potential for enhancing uncertainty estimation. Motivated by their work, we adopt the use of the Beta distribution, a two-dimensional Dirichlet distribution, to model uncertainty in our research and leverage it for further enhancement. Our research takes the initial step towards utilizing uncertainty in an AV system.
\vspace{-1mm}
\section{Dataset}

We utilized the BDD-OIA dataset~\cite{xu2020explainable}, which is an improved rendition of the BDD100K dataset \cite{yu2020bdd100k} and incorporates supplementary annotations for the decision-making process of AVs. The dataset comprises images that are marked with four distinct actions, namely \textit{Forward}, \textit{Stop}, \textit{Turn left}, and \textit{Turn right}, alongside 21 explanatory contexts (See Table \ref{tab:dataset}). For example, if an image portrays a sudden appearance of a pedestrian, it would be labeled as \textit{Stop} due to the explanation of \textit{obstacle: pedestrian}. Furthermore, we performed a thorough cleaning process to eliminate any data noise, including speculating the nonsensical images (images with missing or empty annotations) and removing the duplicated images. It results in a final set of 7,946 training images, 1,118 validation images, and 2,236 testing images. This modified dataset is available upon request.



\begin{table}
\footnotesize
  \caption{BDD-OIA Dataset: Four Action Categories with Corresponding Explanations and Their Frequencies in Dataset}
    \vspace{-3mm}
  \label{tab:dataset}
  \begin{tabular}{p{1.3cm} | c | p{1.5cm}}
    \toprule
    Action & Sample Image & Explanation (\#)\\
    \midrule
    \raisebox{0.8cm}{Move forward}     
    & {\includegraphics[width=0.2\textwidth, height=20mm]{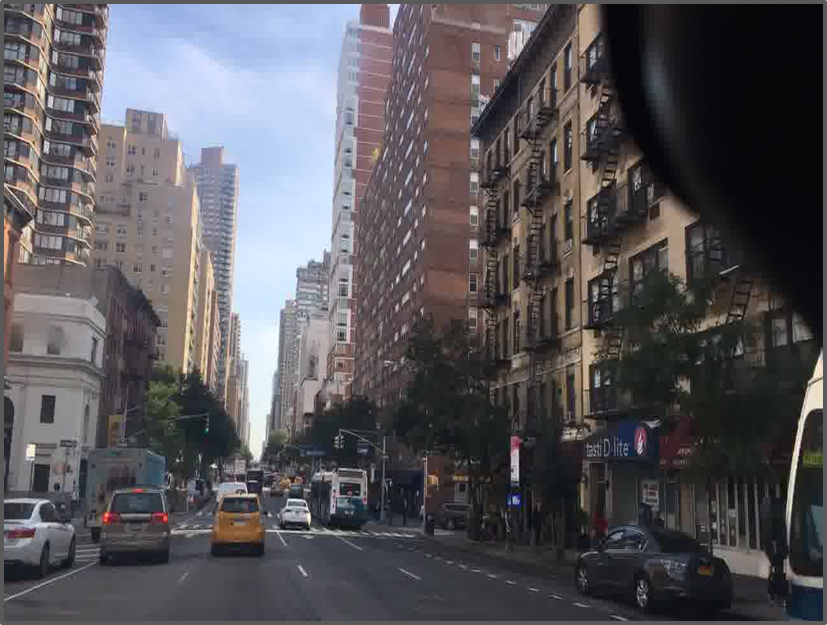}} 
    & \raisebox{0.5cm}{\vbox{\hbox{\strut Follow traffic (2103)}\hbox{\strut Road is clear (1706)}\hbox{\strut Traffic light is green (1105)}}} \\
    \midrule
    \raisebox{0.8cm}{Stop}    
    & {\includegraphics[width=0.2\textwidth, height=20mm]{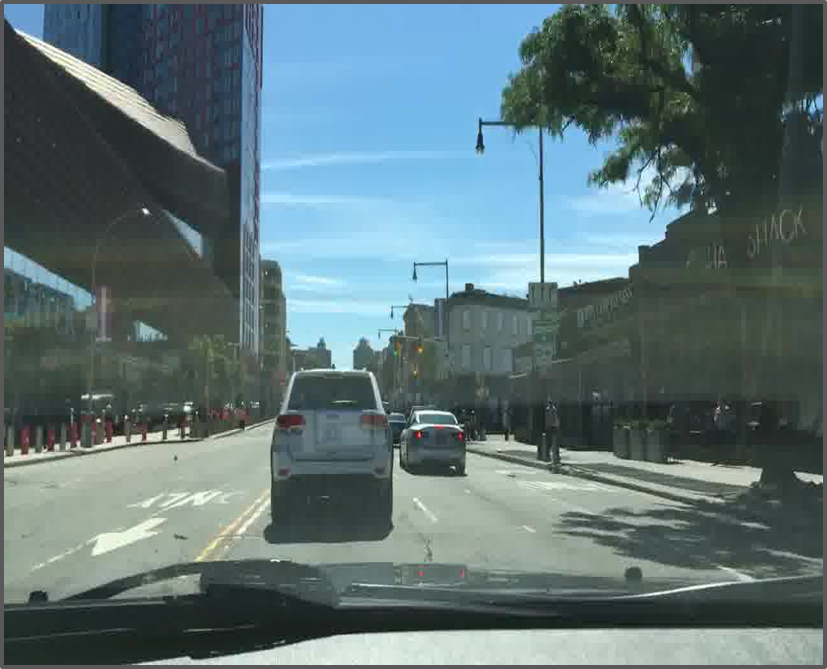}} 
    & \raisebox{0.5cm}{\vbox{\hbox{\strut Obstacles (3076)}\hbox{\strut Traffic light (3648)}\hbox{\strut Traffic sign (383)}}} \\
    \midrule
    \raisebox{0.8cm}{Turn Left}    
    & {\includegraphics[width=0.2\textwidth, height=20mm]{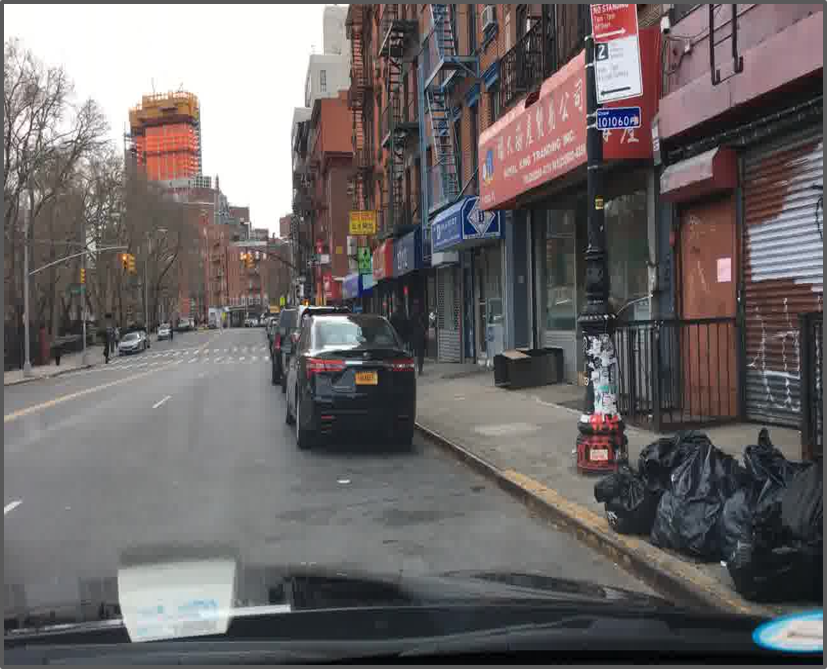}} 
    & \vbox{\hbox{\strut Front car turn left (61)}\hbox{\strut On left-turn lane (260)}\hbox{\strut Traffic allows (226)}\hbox{\strut Obstacles on left (2397)}\hbox{\strut No lane on left (1477)}\hbox{\strut Solid lane on left (2464)}} \\
    \midrule
    \raisebox{0.8cm}{Turn right}    
    & {\includegraphics[width=0.2\textwidth, height=20mm]{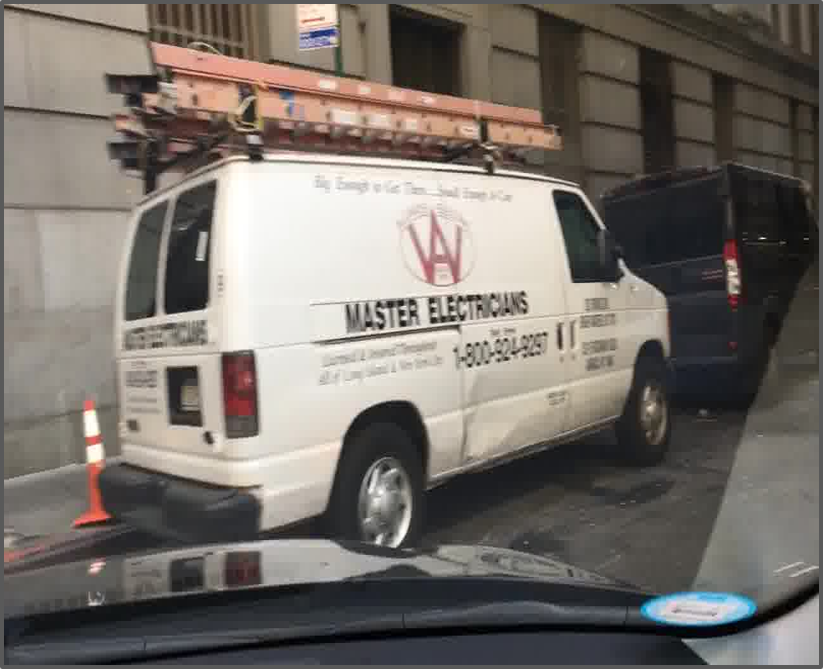}} 
    & \vbox{\hbox{\strut Front car turn right (50)}\hbox{\strut On right-turn lane (452)}\hbox{\strut Traffic allows (226)}\hbox{\strut Obstacles on right (3107)}\hbox{\strut No lane on right (1854)}\hbox{\strut Solid lane on right (1385)}} \\
  \bottomrule
\end{tabular}
\end{table}

\vspace{-1mm}
\section{Methodology}
\begin{figure*}[h]
  \centering
  \includegraphics[width=175mm, scale=1.0]{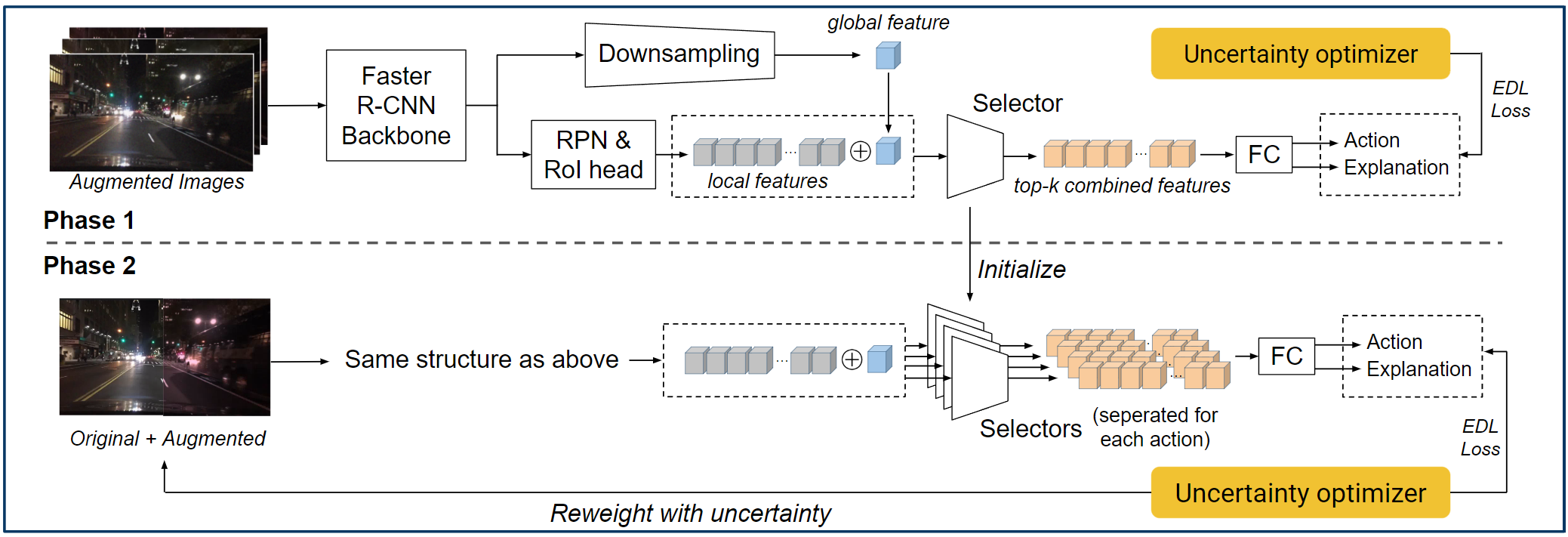}
  \caption{Model Architecture: FC=fully-connected layer, EDL Loss=Evidential Deep Learning Loss Function (Derived using the expected value of the cross-entropy loss over the predicted Beta distribution) }
  \Description{}
  \label{fig:arch}
\end{figure*}

In this section, we explain the concepts underpinning our study
and present an innovative uncertainty-based optimization framework designed for our explainable model. This includes a detailed discussion of our network architecture, the incorporation of uncertainty, and several advanced model training strategies based on uncertainty measurements. Our methodology is composed of two phases: base model construction (Phase 1) and advanced model training (Phase 2). Figure ~\ref{fig:arch} shows the model architecture and training process in two phases. In Phase 1, we concentrate on establishing the foundational architecture and integrating uncertainty, while Phase 2 incorporates further enhancements to the architecture and employs advanced uncertainty-guided training strategies. Detailed explanations of these phases will be provided below.

\vspace{-1mm}
\subsection{Preliminaries}

The foundation of our approach is established on the concept of the explainable model. This type of model has the capability to leverage detected objects within driving scenarios, not merely to predict the corresponding actions of an AV, but also to craft comprehensible explanations. Such explanations are paramount to ensuring users have clarity about the machine's decision-making process.

\vspace{-1mm}
\subsubsection{Object Detection Model.}
In our research, we choose the explainable object-induced model in \cite{xu2020explainable} as a foundational basis to explore and develop our method. A pivotal component of this model is the Faster R-CNN \cite{faster_rcnn} architecture. Originally developed as a real-time object detection system, Faster R-CNN effectively merges the benefits of Region Proposal Networks (RPN) with Fast R-CNN. The RPN component proposes candidate object bounding boxes, while Fast R-CNN uses these boxes to classify the object type. Its backbone, ResNet \cite{he2016deep}, aids in the extraction of global image features. ResNet was specifically architected to train deeper networks by using skip connections or shortcuts to jump over some layers, which solves the vanishing gradient problem in deep neural networks.
\vspace{-1mm}
\subsubsection{Uncertainty Quantification.} 
\label{sec:evidential_uq}
Our proposed method integrates uncertainty quantification through an evidential network, as inspired by the prior work~\cite{sensoy2018evidential}. This approach ventures into the application of the Dirichlet distribution to a rudimentary 10-class classification problem, marking remarkable success in detecting out-of-distribution queries and exhibiting robustness against adversarial perturbations. Similar to the classification problem addressed in the aforementioned study, our chosen explainable object-induced model operates on image features to yield predicted actions and corresponding explanations. 
In addition, the output resembles a collection of independent binary classification outputs. 
Each entry in this collection shows if a particular action or explanation is presented or not. Thus, we opt to utilize Dirichlet distribution as well. The Dirichlet distribution is a probability density function (pdf) for possible values of the probability mass function (pmf) \( p \). It is characterized by \( K \) parameters \( \alpha = [\alpha_1, \ldots, \alpha_K] \) and is given by
\[ D(p | \alpha) = 
\begin{cases}
    \frac{1}{B(\alpha)} \prod_{i=1}^K p_i^{\alpha_i - 1},& \text{if } p \in S_K\\
    0,                                                   & \text{otherwise}
\end{cases}
\]
where \( S_K \) is the K-dimensional unit simplex,
\[ S_K = \left\{ \mathbf{p} \in \mathbb{R}^K : p_i \geq 0, \sum_{i=1}^K p_i = 1 \right\} \]
And \(B(\alpha)\) is the K-dimensional multinomial beta function.
\[ B(\alpha) = \frac{\prod_{i=1}^K \Gamma(\alpha_i)}{\Gamma\left(\sum_{i=1}^K \alpha_i\right)} \]

\vspace{-1mm}
\subsection{Architecture}
Our model leverages both global and local image features. The global features are extracted using the ResNet backbone of the Faster R-CNN. The local features, also referred to as the action-inducing region proposals, are obtained using the Region of Interest (RoI) head in Faster R-CNN. To balance the contribution of both features to the output, the global features are downsampled to match local feature dimensions through convolutional layers and global average pooling. They are then combined into an aggregated set, feeding into the subsequent selector modules. 
The selector module is used to identify the most relevant clues from the aggregated feature set. It consists of multiple convolutional layers, followed by a softmax function. These chosen features are then fed to a stack of fully connected layers, which output the final prediction. 

In this work, we utilize separate selectors for each action. In practice, decisions of different actions are often influenced by distinct visual cues. For example, the ``forward/stop'' action requires more visual focus on the area in front of the vehicle, whereas the ``turn left/right'' action relies more on visual cues from the side view. This approach introduces more interpretability in analyzing the selected regions for reasoning. 

\vspace{-1mm}
\subsection{Uncertainty Integration}

To incorporate uncertainty estimated through the evidential network (Section~\ref{sec:evidential_uq}),   
the initial step is to change the output structure in 
our chosen explainable object-induced model. This object-induced model aims to predict the probabilities of the presence of 4 actions and 21 explanations (See Table \ref{tab:dataset}) and then filters the present actions and explanations by a preset threshold. Now, for each action and explanation, we generate a belief mass distribution. To achieve this, we transform each classification into a belief mass assignment \(\mathbf{b} = <b_1, b_2>\), where the two elements inside represent two statuses: absent or present. Then, the prior distribution for each specific classification of the image is modified to a uniform distribution, i.e., \(Beta(p | <1, 1>)\), a Beta distribution whose parameters are all ones. There is no observed evidence, since the belief masses are all zero. This means that the new classification corresponding to the uniform distribution does not contain any information, and implies total uncertainty. Assume the belief masses become \(\mathbf{b}= <0.6, 0>\) after some training. This means that the total belief in the output is 0.6 and the remaining 0.4 is the uncertainty. The Beta strength is calculated as \(\mathbf{S} = \frac{2}{0.4} = 5\), since \(\mathbf{K} = 2\). Hence, the amount of new evidence derived for the first class is computed as \(5 \times 0.6 = 3\). In this case, the output would correspond to the Beta distribution \(Beta(p | <4, 1>)\). After changing the output format, the next step is to update the loss function. The chosen model originally uses binary cross-entropy with logits loss as its loss function. We change it into a loss function which is derived using the expected value of the cross-entropy loss over the predicted Beta distribution:
\begin{align*}
Li(\Theta) & = \int[\sum_{j=1}^{K} -y_{ij}\log(p_{ij})]\frac{1}{B(\alpha_i)} \prod_{j=1}^K p_{ij}^{\alpha_{ij} - 1} \mathbf{p}_i \\
& = \psi(\log(S_i) - \log(\alpha_{ij})),
\end{align*}
where \(\psi(.)\) is the digamma function. In the rest of this paper, we call the explainable object-induced model that incorporates uncertainty in this way as EDL.

\vspace{-1mm}
\subsection{Uncertainty-Guided Training Strategies}
Based on the uncertainty formulation, we are able to train an uncertainty-aware object-induced explainable model. Specifically, we introduce additional data reweighting and augmentation mechanisms guided by the computed uncertainty.
\vspace{-1mm}
\subsubsection{Data Reweighting.} 
The data reweighting strategy intricately integrates the model's uncertainty metrics. After Phase 1 training, the trained model is used to compute the model and data uncertainties for each image. The model uncertainty is computed by dividing the number of classes (2 for each predicted action or explanation) by the sum of the corresponding Beta distribution. While the data uncertainty is quantified using entropy, specifically through the calculation of the entropy of a given data point, this entropy is obtained by taking the negative summation of the probability of the data point multiplied by the logarithm (base 2) of the same probability. Subsequently, in Phase 2 training, images with uncertainty levels surpassing a threshold are selected for reweighting. The threshold is selected with respect to the highest validation AUC score. Then the training images are reweighed based on this logic: (i) Images for which the model predicts and explains incorrectly or correctly but with high model uncertainty are assigned higher weights. (ii) Images exhibiting excessively high data uncertainty are assigned lower weights because it essentially suggests potential quality issues with these images. The reduced weight serves to minimize their negative impact on model training.
\vspace{-1mm}
\subsubsection{Data Augmentation.} To enhance the model robustness against image noise (e.g., pixel-level noise and different lighting conditions), we incorporate the conventional image augmentation techniques prior to Phase 1 training \cite{shorten2019survey}. The procedure includes brightness and contrast modifications, color adjustments, noise inclusions, and image normalization. However, it is important to note that all the augmentation techniques carry the risk of distorting the genuine context of an image. For example, an excessive color adjustment could lead to a visual semantic shift, turning the red traffic light into purple. To address this concern, we take careful management of how we modify the training images while avoiding drastic alterations. By controlling these changes, we ensure the preservation of the main meaning of the driving scenes. Moving to Phase 2, instead of generating multiple augmented images like in Phase 1, we selectively target the images where our model exhibits heightened uncertainty. This uncertainty is quantified using the same way as the Data reweighting section. Based on this evaluation, we generate the most uncertain variant of the image. Rather than overwhelming the model with numerous variations, we provide it with two critical images: the original and its most uncertain counterpart.
\vspace{-1mm}
\section{Experiment Setup}
\vspace{-1mm}
\subsection{Training Procedures}
Based on the Methodology section, our research builds upon the foundational object-induced action decision model (OIA) proposed in the BDD-OIA publication \cite{xu2020explainable}, while also introducing several components for improvement: (1) incorporating Beta prior into the cross-entropy function (EDL), (2) performing both conventional and uncertainty-guided data augmentation (AG), (3) utilizing separate selectors for each action (SP), and (4) implementing data reweighting (RW). During the initial training phase 1, the EDL base model undergoes conventional training. This foundational phase prioritizes optimal parameter tuning for confident action and explanation predictions. The model processes the labeled dataset and iteratively adjusts weights and biases to minimize the losses derived from the expected value of the cross-entropy loss over the predicted Beta distribution of actions and explanations. To prevent overfitting and promote better generalization, regularization strategies like weight decay and periodic validation checks are incorporated. The robust training conducted in Phase 1 ensures that the model is well-prepared for the next phase. In Phase 2, in addition to the techniques we have incorporated in Phase 1, we refine the model by incorporating the aforementioned components, guided by the prediction uncertainty derived from Phase 1. After that, we evaluate our models, providing a comprehensive analysis of the results, including insights into the contribution of each individual component to the overall performance improvement. 

For more implementation details, we use a batch size of 4 and shuffle the dataset, resizing images to $1280\times736$ pixels. Data augmentation techniques, such as brightness/contrast adjustment, color enhancement, and noise injection, are applied with a 50\% probability for each modification during Phase 1. In Phase 2, these techniques are selectively utilized for uncertain images. The training is conducted using the Adam optimizer for a maximum of 50 epochs and terminates when the validation loss and AUC reach a stable state. In Phase 1, a new model is initialized with a learning rate of 0.001 and weight decay of 0.0001. The learning rate is reduced every 10 epochs. Phase 2 begins by utilizing the parameters of the best model from Phase 1, with a learning rate of 0.00001 and weight decay of 0.000001. The learning rate is adjusted every 5 epochs. It is important to know that the uncertainty-guided data augmentation or reweighting techniques are also iterative in order to adapt to the changing model uncertainty.

\vspace{-1mm}
\subsection{Evaluation Metrics} 

The model's performance underwent evaluation utilizing various metrics, including F1 score, accuracy, precision, and recall. The F1 score, which effectively combines precision and recall, was utilized as the principal metric for evaluating the model's capability to distinguish between true and false predictions. An in-depth analysis of the F1 scores for each action allowed for insights into the model's performance within specific driving scenarios.
\vspace{-1mm}
\section{Results} 
\vspace{-1mm}
\subsection{Performance Metrics}

\begin{table*}
  \caption{Model F1 scores over action predictions and explanation generations (F=Forward, S=Stop, L=Turn Left, R=Turn Right)}
  \vspace{-3mm}
  \label{tab:f1}
  \begin{tabular}{c  c  p{1.5cm}  p{1.5cm}  p{1cm}  p{1cm}  p{1cm}  p{1cm}  p{1cm}}
    \toprule
    \hline
    Phase 1& Phase 2&Actions&Explanations&F&S&L&R\\
    \midrule
    OIA (Baseline1) & - & 0.665 & 0.452 & 0.631 & 0.806 & 0.456 & 0.417\\
    OIA+EDL & - & 0.707 & 0.595 & \textbf{0.686} & \textbf{0.826} & 0.431 & 0.516\\
    OIA+EDL+AG & - & 0.703 & 0.581 & 0.628 & 0.821 & 0.466 & 0.497\\
    \midrule
    OIA+EDL+AG (Baseline2) & - & 0.703 & 0.581 & 0.628 & 0.821 & 0.466 & 0.497\\
    OIA+EDL+AG & SP & 0.707 & 0.599 & 0.672 & 0.824 & 0.474 & 0.484\\
    OIA+EDL+AG & RW & 0.724 & 0.602 & 0.675 & 0.818 & 0.524 & 0.527\\
    \textbf{OIA+EDL+AG} & \textbf{SP+RW} & \textbf{0.727} & \textbf{0.603} & 0.676 & 0.814 & \textbf{0.546} & \textbf{0.544}\\
    OIA+EDL+AG & SP+AG+RW & 0.716 & 0.599 & 0.661 & 0.811 & 0.496 & 0.517\\
    \hline
  \bottomrule
\end{tabular}
\end{table*}

Tables ~\ref{tab:f1} and ~\ref{tab:acc} provide a comprehensive overview of the models' performance across various metrics. The tables are composed of two parts and each of them focuses on a specific phase. Phase 1 (the first three rows of tables) regards the original OIA model as the baseline and compares our proposed uncertainty-integrated models with it. Phase 2 (the rest rows of tables) chooses the best proposed model from Phase 1 as a baseline and compares different enhancement strategies against it. 
\vspace{-1mm}
\subsubsection{F1 Score.}
In Phase 1, the OIA+EDL model performs better than the baseline model (OIA) for overall actions, achieving a significantly higher F1 score of 0.707 compared to the baseline's score of 0.665. For individual actions, the OIA+EDL model shows superior performance in predicting Forward (F) and Stop (S) actions, with F1 scores of 0.686 and 0.826, respectively. For the Left (L) and Right (R) actions, the OIA+EDL+AG model achieves the highest F1 scores of 0.466 and 0.497, respectively. For explanations, the OIA+EDL model once again leads with an impressive F1 score of 0.595, representing a significant improvement from the baseline's score of 0.452.

In Phase 2, when examining the overall F1 scores for actions, the OIA+EDL+AG model with SP+RW configuration demonstrates the highest performance, achieving an F1 score of 0.727. Further exploring individual actions, the OIA+EDL+AG model with SP+RW configuration outperforms in both Left (L) and Right (R) actions, scoring 0.546 and 0.544, respectively. The Forward (F) action is most accurately predicted by the OIA+EDL model, achieving a score of 0.676. In terms of explanations, the OIA+EDL+AG model with SP+RW configuration stands out with an F1 score of 0.603.

\begin{table*}
  \caption{Accuracy, Precision, and Recall of models over action predictions and explanation generations}
    \vspace{-3mm}
  \label{tab:acc}
  \begin{tabular}{c | c | p{1.5cm} | p{1.5cm} | p{1.5cm} | p{1.5cm} | p{1.5cm} | p{1.5cm}}
    \toprule
    \hline
    Phase 1& Phase 2&Actions Accuracy&Explanations Accuracy&Actions Precision&Explanations Precision&Actions Recall&Explanations Recall\\
    \midrule
    OIA (Baseline1) & - & 75.2\% & 90.8\% & 23.3\% & \textbf{71.2}\% & 68.6\% & 40.6\%\\
    OIA+EDL & - & \textbf{79.8}\% & \textbf{91.5}\% & 23.6\% & 65.7\% & 70.4\% & 61.3\%\\
    OIA+EDL+AG & - & 78.3\% & 91.2\% & 23.7\% & 64.7\% & 71.5\% & 60.3\%\\
    \midrule
    OIA+EDL+AG (Baseline2) & - & 78.3\% & 91.2\% & 23.7\% & 64.7\% & 71.5\% & 60.3\%\\
    OIA+EDL+AG & SP & 79.1\% & 91.2\% & 23.8\% & 63.8\% & 71.5\% & 63.5\%\\
    OIA+EDL+AG & RW & 79.3\% & 91.2\% & 23.5\% & 64.0\% & 70.7\% & 63.1\%\\
    OIA+EDL+AG & SP+RW & 77.7\% & 90.6\% & 26.9\% & 60.6\% & 80.0\% & \textbf{67.3}\%\\
    OIA+EDL+AG & SP+AG+RW & 75.7\% & 90.7\% & \textbf{28.3}\% & 61.3\% & \textbf{82.9}\% & 66.2\%\\
    \hline
  \bottomrule
\end{tabular}
\end{table*}
\vspace{-1mm}
\subsubsection{Accuracy.} In the initial phase, the OIA+EDL model demonstrates the highest accuracy for actions at 79.8\% and explanations at 91.5\%. Moving to Phase 2, the OIA+EDL+AG model with RW configuration achieves the highest accuracy for actions at 79.3\%, while for explanations, the accuracy remains consistently high at 91.2\% across models with SP, RW, and even the baseline.
\vspace{-1mm}
\subsubsection{Precision.} In Phase 1, all models demonstrate closely comparable precision scores. Specifically, the OIA+EDL+AG model achieves a precision of 23.7\% for actions and 64.7\% for explanations. In the following phase, the OIA+EDL+AG model combined with SP+RW configuration exhibits the highest precision for actions, reaching 26.9\%. Among explanations, the OIA+EDL+AG model with RW setup attains the highest precision of 64\%.
\vspace{-1mm}
\subsubsection{Recall.} In Phase 1, the OIA+EDL model achieves the highest recall values for actions at 70.4\% and explanations at 61.3\%. Advancing to Phase 2, the OIA+EDL+AG model combined with SP+RW configuration achieves the highest recall for actions at a notable 80\%. Additionally, for explanations, this model excels and yields the top recall of 67.3\%.

\vspace{-1mm}
\subsection{Practical Improvement Demonstrations}
\begin{figure*}[h]
  \centering
  \includegraphics[width=170mm, scale=1.0]{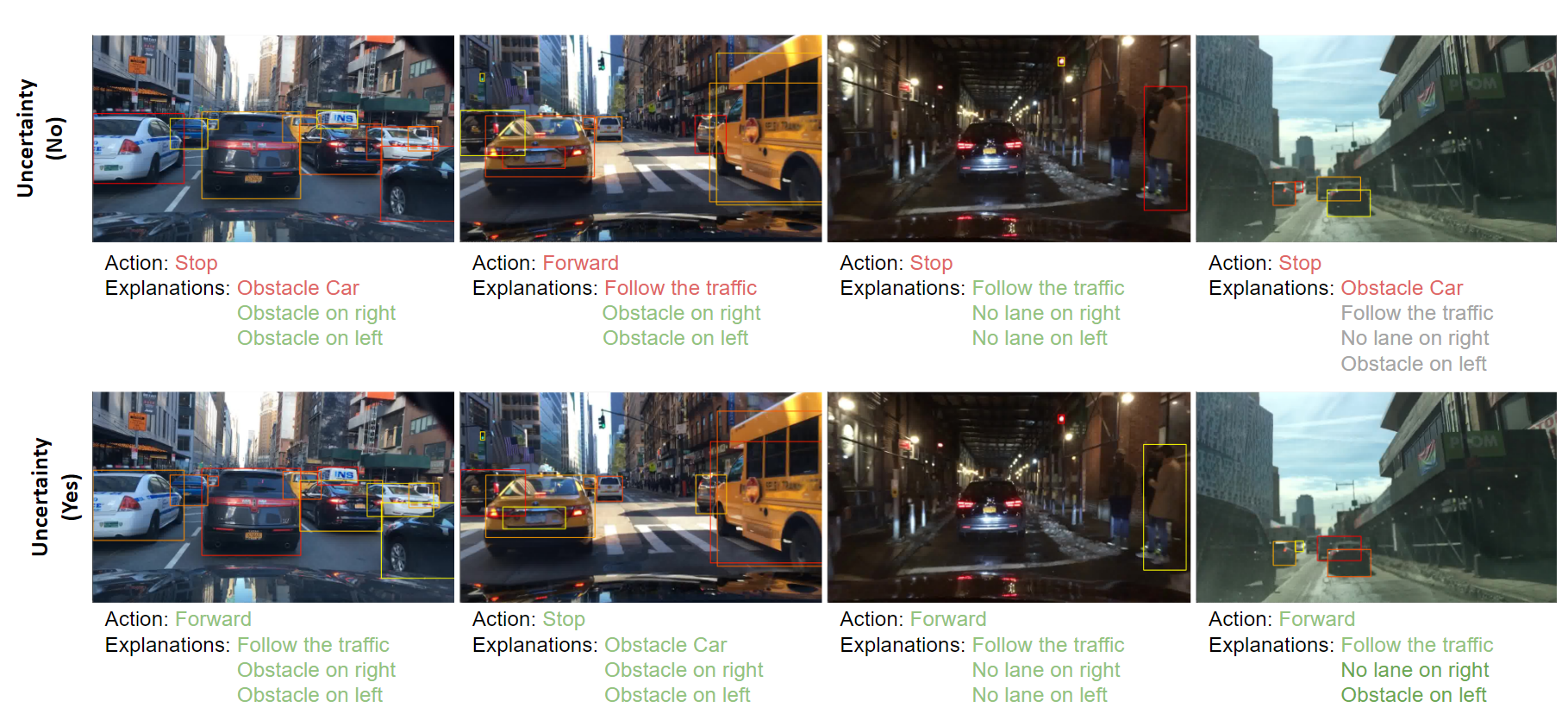}
    \vspace{-4mm}
  \caption{Comparison with Original OIA: the color of bounding boxes reflects the rank of importance (red to yellow = most to least important), and the color of actions and explanations reflects the correctness (green=correct, red=incorrect, gray=missing)}
  \Description{}
  \label{fig:compare-1}
\end{figure*}
\begin{figure*}[h]
  \centering
  \includegraphics[width=160mm, scale=1.0]{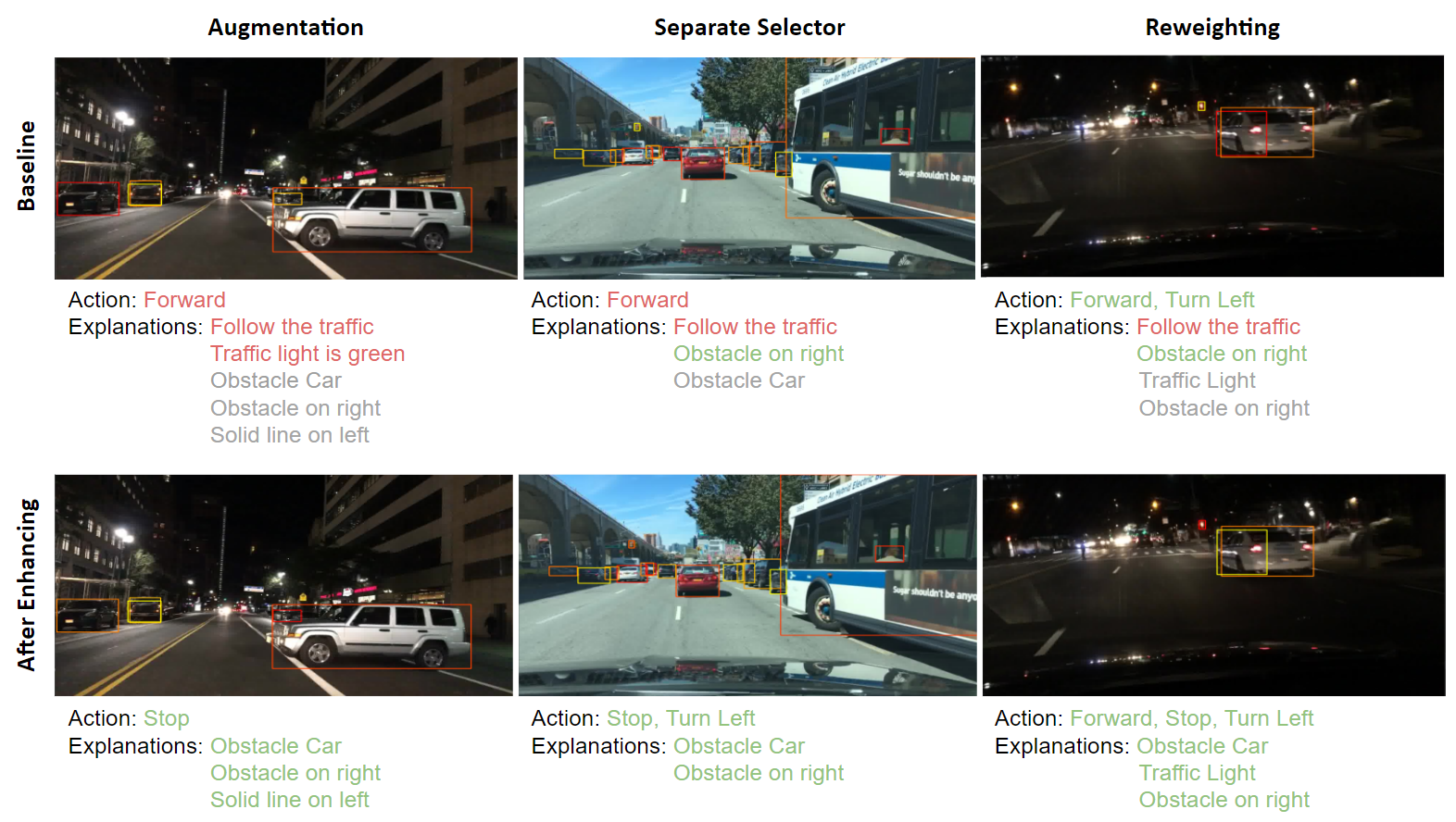}
    \vspace{-4mm}
  \caption{Comparative Efficacy of Enhanced Models: Bounding box colors indicate importance rank (red to yellow = most to least), while action and explanation colors signify correctness (green=correct, red=incorrect, gray=missing)}
  \Description{}
  \label{fig:compare-2}
\end{figure*}
\begin{figure*}[h]
  \centering
  \includegraphics[width=160mm, scale=1.0]{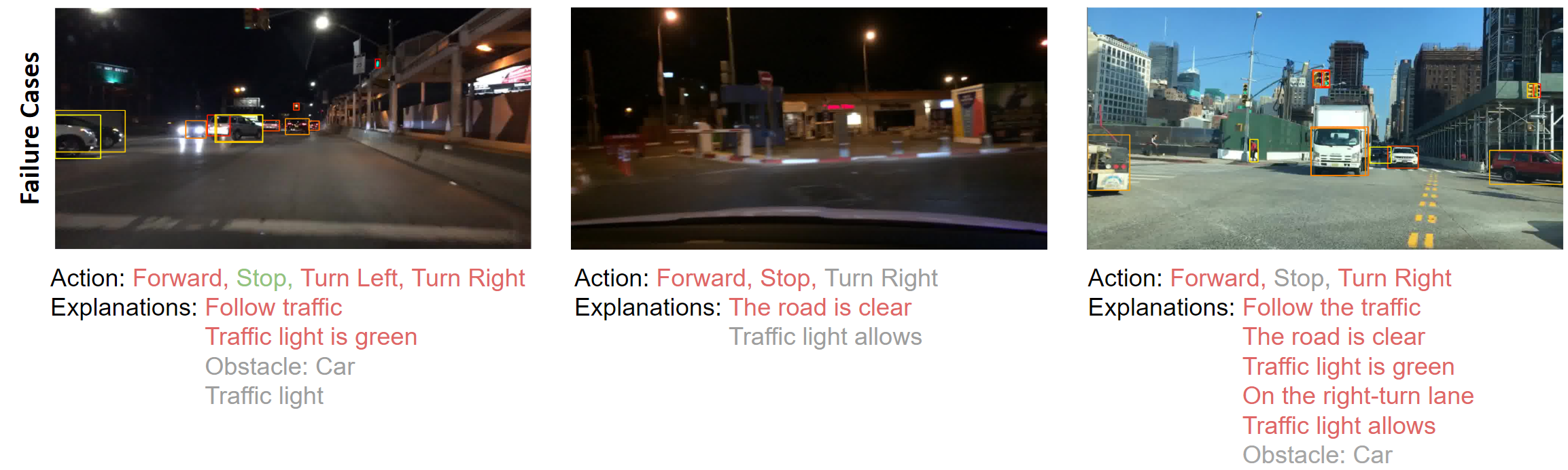}
    \vspace{-4mm}
  \caption{Sample Failure Cases: the color of bounding boxes reflects the rank of importance (red to yellow = most to least important), and the color of actions and explanations reflects the correctness (green=correct, red=incorrect, gray=missing)}
  \Description{}
  \label{fig:fail}
\end{figure*}
Figures ~\ref{fig:compare-1} and ~\ref{fig:compare-2} provide comparative illustrations of real-world scenarios, showcasing the improvement in action prediction and explanation generation achieved through our methods. Each real-world scenario figure includes colored bounding boxes that indicate the importance of selected objects used for decision-making, with color reflecting the rank of importance (red to yellow indicating most to least important). Additionally, each figure presents the corresponding model's actions and explanations. The color is used to indicate correctness, with green representing correct, red indicating incorrect, and gray denoting missing.

Figure ~\ref{fig:compare-1} illustrates the competencies of the enhanced models in comparison to the base OIA model. Our model's superior situational acuity is evident as it demonstrates adeptness in discerning, ranking, and proactively responding to salient elements within scenarios, highlighting its enhanced perceptual abilities.

Figure ~\ref{fig:compare-2} provides a comparison between our base-enhanced model and its subsequent refinements. The advantages of our advanced training strategies are prominently evident across different scenarios. Through data augmentation, the model exhibits an increased capability to correlate salient object recognition with precise predictions and consistent explanations. Additionally, the refined selector demonstrates superior object prioritization, emphasizing the model's enhanced analytical depth. Lastly, the reweighting technique demonstrates its strength in intricate scenarios, further solidifying the model's robust decision-making framework.

\vspace{-1mm}
\section{Discussion}
\vspace{-1mm}
\subsection{Performance Improvement}
In this study, we provide invaluable quantitative results (Tables \ref{tab:f1} \& \ref{tab:acc}) that illuminate the performance metrics of various model configurations. Throughout the initial phase, models that incorporate evidential deep learning (EDL) consistently outperform the baseline model. The OIA+EDL model, in particular, excels in performance across all metrics. This improved performance, evident in the higher F1 scores for overall action and explanation metrics, highlights the benefits of incorporating uncertainty into the model. When transitioning to Phase 2, our results not only show that our uncertainty-guided training strategies (SP \& RW) can bring further improvement to the well-performed OIA+EDL model but also emphasize the effectiveness of combining diverse strategies. For example, the combination of OIA+EDL+AG model with SP+RW configuration achieves the highest F1 scores for both actions and explanations. This illustrates that while incorporating uncertainty brings fundamental improvements, further enhancement can be achieved in the subsequent phase through targeted strategies. The trends in accuracy align with those observed in F1 scores. The OIA+EDL model demonstrates high accuracy metrics in Phase 1, indicating its reliability. Additionally, precision and recall measures provide further support for these findings. The results suggest that models incorporating uncertainty and additional strategies achieve a more balanced and superior performance across various metrics, although the baseline performs well in specific metrics such as explanation precision.

Additionally, we examined a set of images from the test dataset to derive case studies, with a specific focus on instances where the baseline model had difficulties in accurately predicting actions or generating explanations. We present several 
outcomes of our analysis in Figures ~\ref{fig:compare-1} and ~\ref{fig:compare-2}. Figure ~\ref{fig:compare-1} displays the practical improvements achieved by incorporating uncertainty into model training across four real-world scenarios. By comparing the rationality of the ranked importance of selected objects and the accuracy of model results (actions \& explanations), it can be deduced that these improvements are noteworthy. Furthermore, Figure ~\ref{fig:compare-2} demonstrates the practical enhancements obtained by implementing our uncertainty-guided training strategies: In the first scenario, despite both models (baseline and data augmentation-enhanced) selecting the same significant objects in a dark environment, the model with data augmentation enhancement is capable of making accurate actions and providing precise explanations; In the second scenario, the model enhanced with a separate selector 
produces a ranking consistent with the importance of selected objects and can successfully identify that priority should be given to the turning bus on the right side. As a result, it makes more precise decisions than the baseline model; The third scenario presents a challenging environment that allows multiple action choices. The model enhanced with reweighting can determine all suitable actions along with reasonable explanations, surpassing the performance of the baseline model.

In summary, our 
results underscore the benefits of incorporating uncertainty and then optimizing further with targeted strategies subsequently. The consistent improvement across various metrics and strategies reinforces the importance and effectiveness of this two-phase approach. The ability of our model to handle uncertainty robustly holds particular importance for AV systems, considering their exposure to diverse and unforeseen conditions. While our current model is not yet suitable for direct AV application, the underlying principles can guide the enhancement of AV algorithms. This includes the potential integration of multimodal data and `human-in-the-loop' feedback mechanisms for critical situations. Such enhancements have the potential to make AVs more adaptable to complex driving scenarios while adhering to safety requirements.

\vspace{-1mm}
\subsection{Limitation}
While our approach exhibits performance enhancements, limitations persist in the model design and experimental scope. The model's dependence on the Faster R-CNN framework hinders accurate lane distinction in multi-directional traffic scenarios. As a result, the model misidentifies cars from different lanes as being in its lane, resulting in inaccurate predictions and explanations as illustrated in the third case shown in Figure ~\ref{fig:fail}. A potential solution involves integrating advanced lane detection algorithms to improve spatial discernment. Additionally, the complex nature of certain images such as the first and second cases shown in Figure ~\ref{fig:fail} leads to ambiguous annotations, necessitating the adoption of enhanced annotation process to achieve higher accuracy. Another deficiency of the model is its inability to handle images with high data uncertainty. This limitation could be mitigated by integrating dynamic data sources, such as video sequences. Lastly, our findings, derived from a specific dataset and model architecture, possess limited generalizability. Future research should encompass validation on various models and datasets.

\vspace{-1mm}
\subsection{Future Work}


Expanding upon our advancements in uncertainty strategies, we will broaden its applicability across diverse Human-Robot Interaction (HRI) domains. We intend to combine our approach with state-of-the-art multimodal data fusion techniques, integrating multiple data sources to construct a robust and proficient model. This integration will enhance perception and decision-making in high-stakes scenarios. Additionally, we propose training the model not only to handle uncertainty but also to actively engage users during critical moments. By leveraging user input, the model can navigate emergent situations more effectively. We anticipate that this collaborative interaction will refine the model's performance and align it more closely with human intuition and safety priorities.

\vspace{-1mm}
\section{Conclusion}
Our investigation into enhancing the OIA model with uncertainty has been enlightening and valuable. By combining augmentation strategies, a standalone selector, and an uncertainty-driven reweighting mechanism, we have achieved significant improvements in action prediction and rationale formulation. Our model surpasses benchmarks in various scenarios, demonstrating its effectiveness in prediction and reasoning for AV decisions.


\newpage
\bibliographystyle{ACM-Reference-Format}
\balance
\bibliography{Reference/HRI_Paper}

\end{document}